\documentstyle[jair,twoside,11pt,theapa]{article}
\input epsf 

\jairheading{27}{2006}{55-83}{10/05}{09/06}
\ShortHeadings{Minimizing Conflicts: A Heuristic Repair Method}
{Chai, Prasov,  \& Qu}
\firstpageno{55}

\begin{document}

\title{Cognitive Principles in Robust Multimodal Interpretation}

\author{\name Joyce Y. Chai \email jchai@cse.msu.edu \\
       \name Zahar Prasov\email prasovza@cse.msu.edu \\
       \name Shaolin Qu\email qushaoli@cse.msu.edu \\
       \addr Department of Computer Science and Engineering \\
       Michigan State University\\
      East Lansing, MI  48824 USA
 }


\maketitle

\begin{abstract}
Multimodal conversational interfaces provide a natural means for users to communicate with computer systems through multiple modalities such as speech and gesture. To build effective multimodal interfaces, automated interpretation of user multimodal inputs is important. 
Inspired by the previous investigation on cognitive status in multimodal human machine interaction, we have developed a greedy algorithm for interpreting user referring expressions (i.e., multimodal reference resolution). This algorithm incorporates the cognitive principles of Conversational Implicature and Givenness Hierarchy and applies constraints from various sources (e.g., temporal, semantic, and contextual) to resolve references.
Our empirical results have shown the advantage of this algorithm in efficiently resolving a variety of user references.  
Because of its simplicity and generality, this approach has the potential to improve the robustness of multimodal input interpretation.

\end{abstract}

\section{Introduction}
\label{Introduction}

Multimodal systems provide a natural and effective way for users to interact with computers through multiple modalities such as speech, gesture, and gaze. Since the first appearance of the ``Put-That-There" system \cite{bolt:80}, a number of multimodal systems have been built, among which there are systems that combine speech, pointing \cite{neal:91,stock:93}, and gaze \cite{koons:93}, systems that integrate speech with pen inputs (e.g., drawn graphics) \cite{cohen:96,wahlster:98}, systems that combine multimodal inputs and outputs \cite{cassell:90}, systems in mobile environments \cite{oviatt:00}, and systems that engage users in an intelligent conversation \cite{gustafson:00,stent:99}. Earlier studies have shown that multimodal interfaces enable users to interact with computers naturally and effectively \cite{oviatt:96,oviatt:99}.

One important aspect of building multimodal systems is multimodal interpretation, which is a process that identifies the meanings of user inputs. In particular, a key element in multimodal interpretation is known as reference resolution, which is a process that finds the most proper referents to referring expressions. Here a referring expression is a phrase that is given by a user in her inputs (most likely in speech inputs) to refer to a specific entity or entities. A referent is an entity (e.g., a specific object) to which the user refers. Suppose that a user points to \textsf{House 6} on the screen and says {\em how much is this one}. In this case, reference resolution must infer that the referent \textsf{House 6} should be assigned to the referring expression {\em this one}. This paper particularly addresses this problem of reference resolution in multimodal interpretation.

In a multimodal conversation, the way users communicate with a system depends on the available interaction channels and the situated context (e.g., conversation focus, visual feedback). These dependencies form a rich set of constraints from various aspects (e.g., semantic, temporal, and contextual). A correct interpretation can only be attained by simultaneously considering these constraints. 

Previous studies have shown that user referring behavior during multimodal conversation does not occur randomly, but rather follows certain linguistic and cognitive principles.  In human machine interaction, earlier work has shown strong correlations between the cognitive status in Givenness Hierarchy and the form of referring expressions \cite{kehler:00}.
Inspired by this early work, we have developed a greedy algorithm for multimodal reference resolution. This algorithm incorporates the principles of Conversational Implicature and Givenness Hierarchy and applies constraints from various sources (e.g., gesture, conversation context, and visual display).
Our empirical results have shown the promise of this algorithm in efficiently resolving a variety of user references. One major advantage of this greedy algorithm is that the prior linguistic and cognitive knowledge can be used to guide the search and prune the search space during constraint satisfaction. Because of its simplicity and generality, this approach has the potential to improve the robustness of interpretation and provide a practical solution to multimodal reference resolution \cite{chai:05}.

In the following sections, we will first demonstrate different types of referring behavior observed in our studies.  We then briefly introduce the underlying cognitive principles for human-human communication and describe how these principles can be used in a computational model to efficiently resolve multimodal references. Finally, we will present the experimental results.

\section{Multimodal Reference Resolution}
\label{refresolution}

\begin{figure}
\begin{center}
\epsfxsize=3.5in {\epsfbox{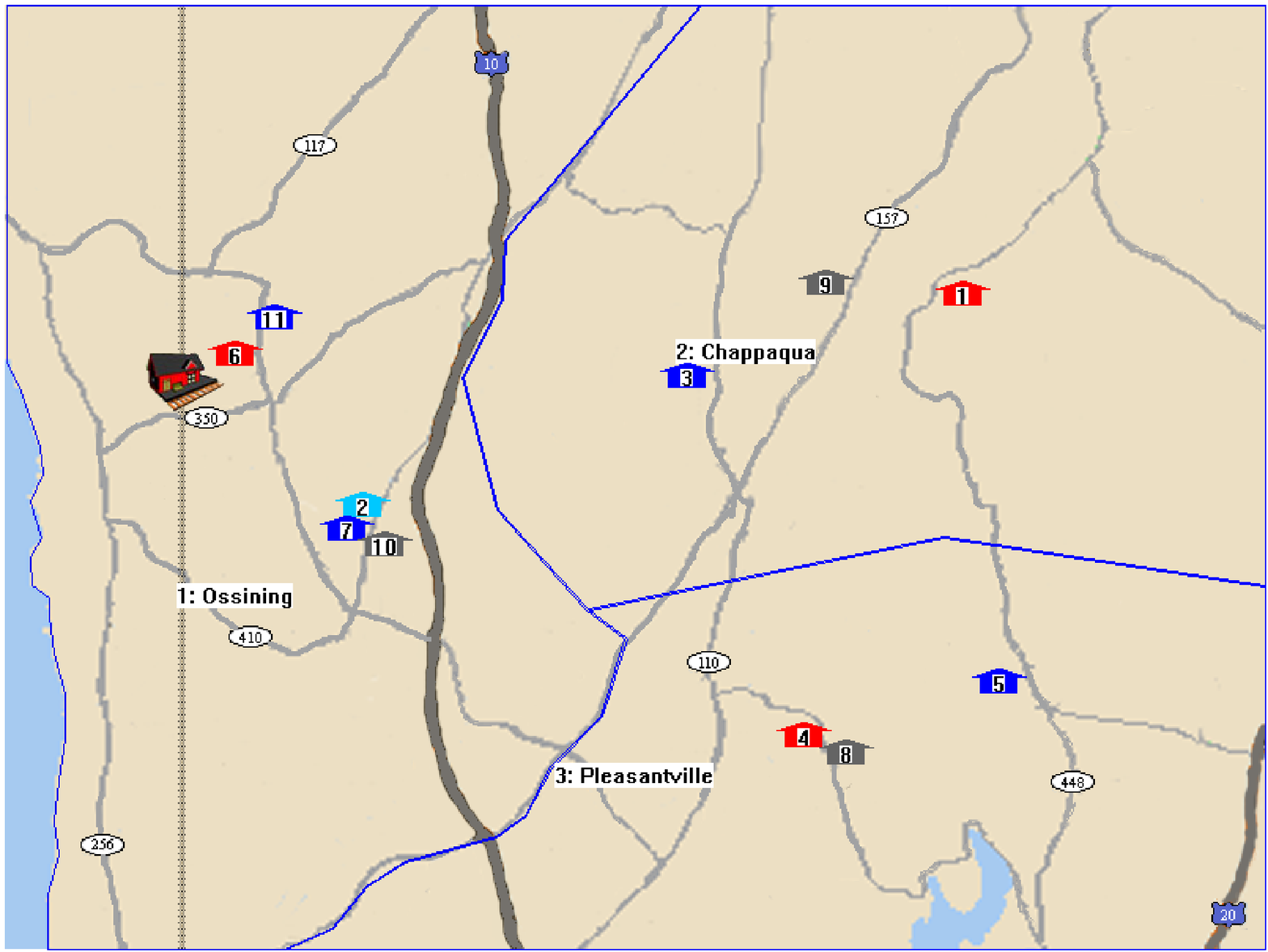}}
\caption{\small A snapshot of the multimodal conversational system.}
\label{fig:snapshot}
\end{center}
\end{figure}

In our previous work \cite{chai:04a,chai:04c}, a multimodal conversational system was developed for users to acquire real estate information\footnote{The first prototype of this system was developed at IBM T. J. Watson Research Center with P. Hong, M. Zhou, and colleagues at the Intelligent Multimedia Interaction group.}. Figure~\ref{fig:snapshot} is the snapshot of a graphical user interface. Users can interact with this interface through both speech and gesture. Table~\ref{tab:fragment} shows a fragment of the conversation.

\begin{table}[t]
\center
\begin{tabular}{|l|l|} \hline
  $U_1$  & {\em Speech:} How much does this cost? \\
         & {\em Gesture:} point to a position on the screen      \\      \hline
$S_1$   &  {\em Speech:} The price is 400K \\
              &  {\em Graphics:} highlight the house in discussion   \\  \hline
$U_2$   & {\em Speech:} How large? \\ \hline
$S_2$   & {\em Speech:} 2500 square feet \\ \hline
$U_3$   & {\em Speech:} Compare it with this house and this one \\
            & {\em Gesture:} ....circle....cirle (put two consecutive circles on the screen) \\ \hline
$S_3$   & {\em Speech:} Here are your comparison results \\
        & {\em Graphics:} show a table of comparison  \\ \hline \hline
\end{tabular}
\caption{A fragment demonstrating interaction with different types of referring behavior }
\label{tab:fragment}
\end{table}

In this fragment, the user exhibits different types of referring behavior. For example, the input from $U_1$ is considered as a simple input. This type of simple input only has one referring expression in the spoken utterance and one accompanying gesture. Multimodal fusion that combines information from speech and gesture will likely resolve what {\em this} refers to. In the second user input ($U_2$), there is no accompanying gesture and no referring expression is explicitly used in the speech utterance. At this time, the system needs to use the conversation context to infer that the object of interest is the house mentioned in the previous turn of the conversation. In the third user input, there are multiple referring expressions and multiple gestures. These types of inputs are considered complex inputs. Complex inputs are more difficult to resolve. We need to consider the temporal relations between the referring expressions and the gestures, the semantic constraints specified by the referring expressions, and the contextual constraints from the prior conversation. For example, in the case of $U_3$, the system needs to understand that {\em it} refers to the house that was the focus of the previous turn; and {\em this house} and {\em this one} should be aligned with the two consecutive gestures. Any subtle variations in any of the constraints, including the temporal ordering, the semantic compatibility, and the gesture recognition results will lead to different interpretations.

From this example, we can see that in a multimodal conversation, the way a user interacts with a system is dependent not only on the available input channels (e.g., speech and gesture), but also upon his/her conversation goals, the state of the conversation, and the multimedia feedback from the system. In other words, there is a rich context that involves dependencies from many different aspects established during the interaction. Interpreting user inputs can only be situated in this rich context. For example, the temporal relations between speech and gesture are important criteria that determine how the information from these two modalities can be combined. The focus of attention from the prior conversation shapes how users refer to those objects, and thus, influences the interpretation of referring expressions. Therefore, we need to simultaneously consider the temporal relations between the referring expressions and the gestures, the semantic constraints specified by the referring expressions, and the contextual constraints from the prior conversation. In this paper, we present an efficient approach that is driven by cognitive principles to combine temporal, semantic, and contextual constraints for multimodal reference resolution.

\section{Related Work}
\label{relatedwork}

Considerable effort has been devoted to studying user multimodal behavior \cite{cohen:84,oviatt:00} and mechanisms to interpret user multimodal inputs \cite{chai:04a,gustafson:00,huls:95,johnston:97,johnston:98,johnston:00,kehler:00,koons:93,neal:91,oviatt:97,stent:99,stock:93,wahlster:98,wu:99,zancanaro:97}. 

For multimodal reference resolution, some early work keeps track of a focus space from the dialog~\cite{grosz:86} and a display model to capture all objects visible on the graphical display~\cite{neal:98}. It then checks semantic constraints such as the type of the candidate objects being referenced and their properties  for reference resolution. 
A modified centering model for multimodal reference resolution is also introduced in previous work~\cite{zancanaro:97}. The idea is that based on the centering movement between turns, segments of discourse can be constructed. The discourse entities appearing in the segment that is accessible to the current turn can be used to constrain the referents to referring expressions.  
Another approach is introduced to use contextual factors for multimodal reference resolution~\cite{huls:95}. In this approach, a salience value is assigned to each instance based on the contextual factors. 
To determine the referents of multimodal referring expressions, this approach retrieves the most salient referent that satisfies the semantic restrictions of the referring expressions.  All these earlier approaches have some greedy nature, which is largely dependent on semantic constraints and/or constraints from conversation context.

To resolve multimodal references, there are two important issues. First it is the mechanism to combine information from various sources and modalities.  The second is the capability to obtain the best interpretation (among all the possible alternatives) given a set of temporal, semantic, and contextual constraints. 
In this section, we give a brief introduction to three recent approaches that address these issues. 

\subsection{Multimodal Fusion}
\label{sec:fst}
Approaches to multimodal fusion \cite{johnston:98,johnston:00}, although they focus on a different problem of overall input interpretation, provide effective solutions to reference resolution.
There are two major approaches to multimodal fusion:  unification-based approaches \cite{johnston:98} and finite state approaches \cite{johnston:00}. 

The unification-based approach identifies referents to referring expressions by unifying feature structures generated from speech utterances and gestures using a multimodal grammar \cite{johnston:97,johnston:98}. The multimodal grammar combines both temporal and spatial constraints. Temporal constraints encode the absolute temporal relations between speech and gesture~\cite{johnston:98},.  The grammar rules are predefined based on empirical studies of multimodal interaction \cite{oviatt:97}. For example, one rule indicates that speech and gesture can be combined only when the speech either overlaps with gesture or follows the gesture within a certain time frame. The unification approach can also process certain complex cases (as long as they satisfy the predefined multimodal grammar) in which a speech utterance is accompanied by more than one gesture of different types \cite{johnston:98}. Using this approach to accommodate various situations such as those described in Figure~\ref{fig:snapshot} will require adding different rules to cope with each situation. If a specific user referring behavior did not exactly match any existing integration rules (e.g., temporal relations), the unification would fail and therefore references would not be resolved.

The finite state approach applies finite-state transducers for multimodal parsing and understanding \cite{johnston:00}.  Unlike the unification-based approach with chart parsing that is subject to significant computational complexity concerns~\cite{johnston:00}, the finite state approach provides more efficient, tight-coupling of multimodal understanding with speech recognition. In this approach, a multimodal context-free grammar is defined to transform the syntax of multimodal inputs to the semantic meanings. The domain-specific semantics are directly encoded in the grammar. Based on these grammars, multi-tape finite state automata can be constructed. These automata are used for identifying semantics of combined inputs. Rather than absolute temporal constraints as in the unification-based approach, this approach relies on temporal order between different modalities. During the parsing stage, the gesture input from the gesture tape (e.g., pointing to a particular person) that can be combined with the speech expression in the speech tape (e.g., {\em this person}) is considered as the referent to the expression. A problem with this approach is that the multi-tape structure only takes input from speech and gesture and does not incorporate the conversation history into consideration.

\subsection{Decision List}
\label{sec:dlist}
To identify potential referents,  previous work has investigated Givenness Hierarchy (to be introduced later) in multimodal interaction \cite{kehler:00}.  Based on data collected from Wizard of Oz experiments, this investigation suggests that users tend to tailor their expressions to what they perceive to be the system's beliefs concerning the cognitive status of referents from their prominence (e.g., highlight) on the display. The tailored referring expressions can then be resolved with a high accuracy based on the following decision list:
\begin{enumerate}
\item If an object is gestured to, choose that object.
\item Otherwise, if the currently selected object meets all semantic type constraints imposed by the referring expression, choose that object.
\item Otherwise, if there is a visible object that is semantically compatible, then choose that object.
\item Otherwise, a full NP (such as a proper name) is used to uniquely identify the referent.
\end{enumerate}

From our studies \cite{chai:04b}, we found this decision list has the following limitations:
\begin{itemize}
\item Depending on the interface design, ambiguities (from a system's perspective) could occur. For example, given an interface where one object (e.g., house) can  sometimes be created on top of another object (e.g., town), a pointing gesture could result in multiple potential objects. Furthermore, given an interface with crowded objects, a finger point could also result in multiple objects with different probabilities. The decision list is not able to handle these ambiguous cases.
\item User inputs are not always simple (consisting of no more than one referring expression and one gesture as indicated in the decision list). In fact, in our study \cite{chai:04b}, we found that user inputs can also be complex, consisting of multiple referring expressions and/or multiple gestures. The referents to these referring expressions could come from different sources, such as gesture inputs and conversation context. The temporal alignment between speech and gesture is also important in determining the correct referent for a given expression. The decision list is not able to handle these types of complex inputs.
\end{itemize}
Nevertheless, the previous findings \cite{kehler:00} have inspired this work and provided a basis for the algorithm described in this paper.

\subsection{Optimization}
\label{sec:graphmatching}

\begin{figure}
\begin{center}
\epsfxsize=6in {\epsfbox{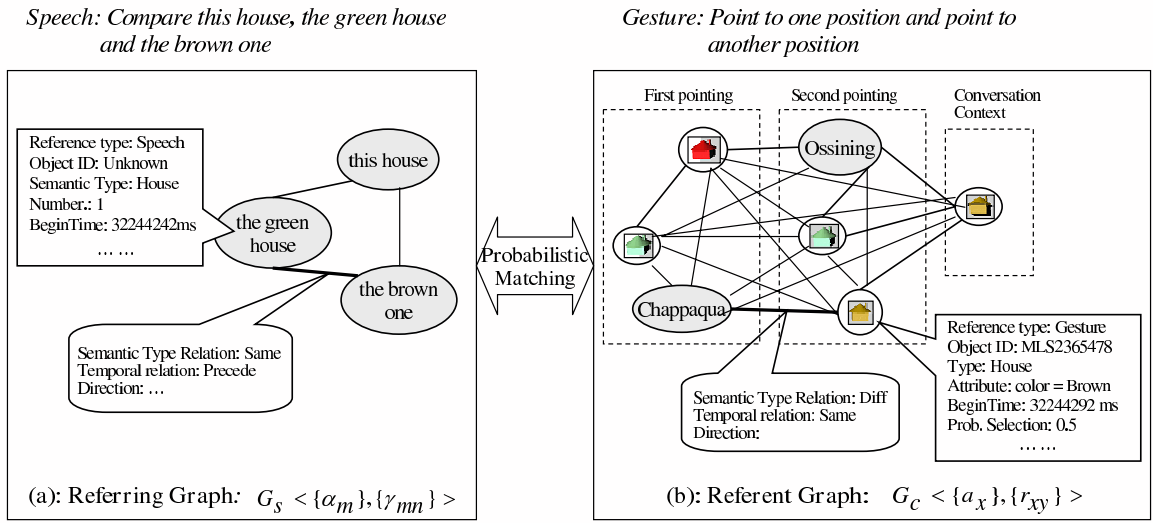}}
\caption{\small Reference resolution through probabilistic graph-matching}
\label{fig:graph}
\end{center}
\end{figure}

Recently, a probabilistic approach was developed for optimizing reference resolution based on graph matching~\cite{chai:04a}. In the graph-matching approach, information gathered from multiple input modalities and the conversation context is represented as attributed relational graphs (ARGs) \cite{tsai:79}. Specifically, two graphs are used. One graph represents referring expressions from speech utterances (i.e., called referring graph). A referring graph contains referring expressions used in a speech utterance and the relations between these expressions. Each node corresponds to one referring expression and consists of the semantic and temporal information extracted from that expression.  Each edge represents the semantic and temporal relation between two referring expressions. The resulting graph is a fully connected, undirected, graph. For example, as shown in Figure~\ref{fig:graph}(a), from the speech input {\em compare this house, the green house, and the brown one}, three nodes are generated in the referring graph representing three referring expressions. Each node contains semantic and temporal features related to its corresponding referring expression. These include the expression's semantic type (house, town, etc.), number of potential referents, type dependent features (size, price, etc.), syntactic category of the expression, and the timestamp of when the expression was produced. Each edge contains features describing semantic and temporal relations between a pair of nodes. The semantic features simply indicate whether or not two nodes share the same semantic type if this can be inferred from the utterance. Otherwise, the semantic type relation is deemed to be unknown. The temporal features indicate which of the two expressions was uttered first.

Similarly, another graph represents all potential referents gathered from gestures, history, and the visual display (i.e., called referent graph). Each node in a referent graph captures the semantic and temporal information about a potential referent, together with its selection probability. The selection probability is particularly applied to objects indicated by a gesture. Because a gesture such as a pointing or a circle can potentially introduce ambiguity in terms of the intended referents, a selection probability is used to indicate how likely it is that an object is selected by a particular gesture. This selection probability is derived by a function of the distance between the location of the entity and the focus point of the recognized gesture on the display.  As in a referring graph, each edge in a referent graph captures the semantic and temporal relations between two potential referents such as whether the two referents share the same semantic type and the temporal order between two referents as they are introduced into the discourse. For example, since the gesture input consists of two pointings, the referent graph (Figure ~\ref{fig:graph}b) consists of all potential referents from these two pointings. The objects in the first dashed rectangle are potential referents selected by the first pointing, and those in the second dashed rectangle correspond to the second pointing. Furthermore, the salient objects from the prior conversation are also included in the referent graph since they could be the potential referents as well (e.g., the rightmost dashed rectangle in Figure ~\ref{fig:graph}b).

Given these graph representations, the reference resolution problem becomes a probabilistic graph-matching problem \cite{gold:96}. The goal is to find a match between the referring graph $G_{s}$ and the referent graph $G_{c}$~\footnote{The subscription $s$ in $G_{s}$ refers to speech referring expressions and $c$ in $G_{c}$ refers to candidate referents.} that achieves the maximum compatibility (i.e., maximizes $Q(G_c, G_s)$) as described in the following equation:

\begin{equation}
\begin{array}{ll}
    Q(G_c, G_s)= &  \sum_x \sum_m P(\alpha_x, \alpha_m) NodeSim(\alpha_x, \alpha_m) \\
                 & + \sum_x \sum_y \sum_m \sum_n  P(\alpha_x, \alpha_m) P(\alpha_y, \alpha_n) EdgeSim(\gamma_{xy}, \gamma_{mn})
                 \end{array}
\end{equation}


$P(\alpha_x, \alpha_m)$ is the matching probability between a referent node $\alpha_x$ and a referring node $\alpha_m$. The overall compatibility $Q(G_c,G_s)$ depends on the node compatibility $NodeSim$ and the edge compatibility $EdgeSim$, which were further defined by temporal and semantic constraints~\cite{chai:04c}. When the algorithm converges, $P(\alpha_x, \alpha_m)$ gives the matching probabilities between a referent node $\alpha_x$ and a referring node $\alpha_m$ that maximizes the overall compatibility function. Using these matching probabilities, the system is able to identify the most probable referent $\alpha_x$ to each referring node  $\alpha_m$. Specifically, the referring expression that matches a potential referent is assigned to the referent if the probability of this match exceeds an empirically computed threshold. If this threshold is not met, the referring expression remains unresolved.  

Theoretically, this approach provides a solution that maximizes the overall satisfaction of semantic, temporal, and contextual constraints. However, like many other optimization approaches, this algorithm is non-polynomial. It relies on an expensive matching process, which attempts every possible assignment, in order to converge on an optimal interpretation based on those constraints. However, previous linguistic and cognitive studies indicate that user language behavior does not occur randomly, but rather follows certain cognitive principles. Therefore, a question arises whether any knowledge from these cognitive principles can be used to guide this matching process and reduce the complexity.

\section{Cognitive Principles}

Motivated by previous work~\cite{kehler:00}, we specifically focus on two principles: Conversational Implicature and Givenness Hierarchy.

\subsection{Conversational Implicature}

Grice's Conversational Implicature Theory indicates that the interpretation and inference of an utterance during communication is guided by a set of four maxims \cite{grice:75}. Among these four maxims, the Maxim of Quantity and the Maxim of Manner are particularly useful for our purpose.

The Maxim of Quantity has two components: (1) make your contribution as informative as is required (for the current purposes of the exchange), and (2) do not make your contribution more informative than is required. In the context of multimodal conversation, this maxim indicates that users generally will not make any unnecessary gestures or speech utterances.  This is especially true for pen-based gestures since they usually require a special effort from a user. Therefore, when a pen-based gesture is intentionally delivered by a user, the information conveyed is often a crucial component used in interpretation.

Grice's Maxim of Manner has four components: (1) avoid obscurity of expression, (2) avoid ambiguity, (3) be brief, and (4) be orderly. This maxim indicates that users will not intentionally make ambiguous references. They will use expressions (either speech or gesture) they believe can uniquely describe the object of interest so that listeners (in this case a computer system) can understand. The expressions they choose depend on the information in their mental models about the current state of the conversation. However, the information in a user's mental model might be different from the information the system possesses. When such an information gap happens, different ambiguities could occur from the system point of view. In fact, most ambiguities are not intentionally caused by the human speakers, but rather by the system's incapability of choosing among alternatives given incomplete knowledge representation, limited capability of contextual inference, and other factors (e.g., interface design issues). Therefore, the system should not anticipate deliberate ambiguities from users (e.g., a user only utters {\em a house} to refer to a particular house on the screen), but rather should focus on dealing with the types of ambiguities caused by the system's limitations (e.g., gesture ambiguity due to the interface design or speech ambiguity due to incorrect recognition).

These two maxims help positioning the role of gestures in reference resolution. In particular, these maxims have put the potential referents indicated by a gesture at a very important position, which is described in Section~\ref{sec:algorithm}.

\subsection{Givenness Hierarchy}

\begin{figure}
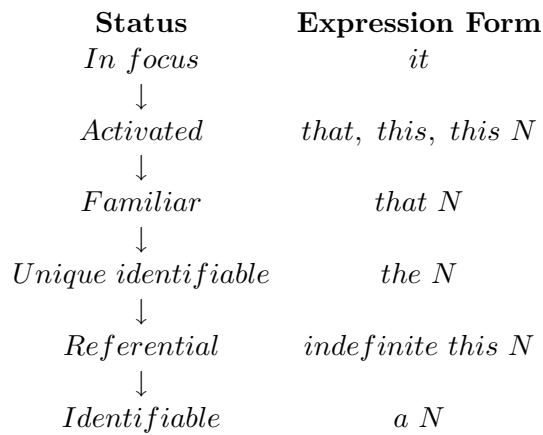

\begin{center}
\( \begin{array}{cc}
{\bf Status} & {\bf Expression\  Form} \\
In\ focus & it \\
\downarrow & \\
Activated & that,\ this,\ this\ N \\
\downarrow & \\
Familiar & that\ N \\
\downarrow & \\
Unique\ identifiable & the\ N \\
\downarrow & \\
Referential  & indefinite\ this\ N \\
\downarrow & \\
Identifiable & a\ N

\end{array}
\)
\caption{\small Givenness Hierarchy}
\label{fig:hierarchy}
\end{center}
\end{figure}

The Givenness Hierarchy proposed by Gundel et al. explains how different determiners and pronominal forms signal different information about memory and attention state (i.e., cognitive status) \cite{gundel:93}.  As in Figure~\ref{fig:hierarchy}, there are six cognitive statuses in the hierarchy. For example, \textsf{In focus} indicates the highest attentional state that is likely to continue to be the topic. \textsf{Activated} indicates entities in short term memory.    Each of these statuses is associated with some forms of referring expressions.  In this hierarchy, each cognitive status implies the statuses down the list. For example, \textsf{In focus} implies \textsf{Activated}, \textsf{Familiar}, etc.  The use of a particular expression form not only signals that the associated cognitive status is met, but also signals that all lower statuses have been met. In other words, a given form that is used to describe a lower status can also be used to refer to a higher status, but not vice versa. Cognitive statuses are necessary conditions for appropriate use of different forms of referring expressions. Gundel et al. found that different referring expressions almost exclusively correlate with the six statuses in this hierarchy.

The Givenness Hierarchy has been investigated earlier in algorithms for resolving pronouns and demonstratives in spoken dialog systems~\cite{eckert:00,byron:02} and in multimodal interaction \cite{kehler:00}. In particular, we would like to extend the previous work~\cite{kehler:00} and investigate whether Conversational Implicature and Givenness Hierarchy can be used to resolve a variety of references from simple to complex, and from precise to ambiguous. Furthermore, the decision list used in Kehler (2000) is proposed based on data analysis and has not been implemented or evaluated in a real-time system. Therefore, our second goal is to design and implement an efficient algorithm by incorporating these cognitive principles and empirically compare its performance with the optimization approach~\cite{chai:04c}, the finite state approach~\cite{johnston:00}, and the decision list approach~\cite{kehler:00}.  

\section{A Greedy Algorithm}
\label{sec:algorithm}

A greedy algorithm always makes the choice that looks best at the moment of processing. That is, it makes a locally optimal choice in the hope that this choice will lead to a globally optimal solution. Simple and efficient greedy algorithms can be used to approximate many optimization problems. Here we explore the use of Conversational Implicature and Givenness Hierarchy in designing an efficient greedy algorithm. In particular, we extend the decision list from Kehler (2000) and utilize the concepts from the two cognitive principles in the following way:
\begin{itemize}
\item Corresponding to the Givenness Hierarchy, the following hierarchy holds for potential referents: $Focus > Visible$. This hierarchy indicates that objects in focus have higher status in terms of attention states than objects in the visual display. Here {\em Focus} corresponds to the cognitive statuses \textsf{In focus}  and \textsf{Activated} in the Givenness Hierarchy, and {\em Visible} corresponds to the statuses \textsf{Familiar} and \textsf{Uniquely identifiable}. Note that Givenness Hierarchy is fine grained in terms of different statuses. Our application may not be able to distinguish the difference between these statuses (e.g., \textsf{In focus}  and \textsf{Activated}) and effectively use them. Therefore, {\em Focus} and {\em Visible} are introduced here to group some similar statuses (with respect to our application) together. Since there is a need to differentiate the objects that have been mentioned recently (e.g., in focus and activated) and objects that are accessible either on the graph display or from the domain model (e.g., familiar and unique identifiable), we assign them to different modified statuses (e.g., {\em Focus} and {\em Visible}). 

\item   Based on the Conversational Implicature, since a pen-based gesture takes a special effort to deliver, it must convey certain useful information. In fact, objects indicated by a gesture should have the highest attentional state since they are deliberately singled out by a user.
Therefore, by combining (1) and (2), we derive a modified hierarchy $Gesture > Focus > Visible > Others$. Here {\em Others} corresponds to indefinite cases in Givenness Hierarchy. This modified hierarchy coincides with the processing order of the Kehler's decision list (2000). This modified hierarchy will guide the greedy algorithm in its search for solutions.  Next, we describe in detail the algorithm and related representations and functions.
\end{itemize}

\subsection{Representation}
\label{sec:representation}

At each turn\footnote{Currently, user inactivity (i.e., 2 seconds with no input from either speech or gesture) is used as the boundary to decide an interaction turn.} (i.e., after receiving a user input) of the conversation, we use three vectors to represent the first three statuses in our modified hierarchy: objects selected by a gesture, objects in the focus, and objects visible on the display as follows:
\begin{itemize}

\item Gesture vector ($\vec{g}$) captures objects selected by a series of gestures. Each element $g_{i}$ is an object potentially selected by a gesture. For elements $g_{i}$  and $g_{j}$  where $i < j$, the gesture that selects objects $g_{i}$  should: 1) temporally precede  the gesture that selects $g_{j}$ or 2) be the same as the gesture that selects $g_{j}$ since one gesture could result in multiple objects.
\item Focus vector ($\vec{f}$) captures objects that are in the focus but are not selected by any gesture. Each element represents an object considered to be the focus of attention from the previous turn of the conversation. There is no temporal precedence relation between these elements. We consider all the corresponding objects are simultaneously accessible to the current turn of the conversation.
\item Display vector ($\vec{d}$) captures objects that are visible on the display but are neither selected by any gesture (i.e., $\vec{g}$) nor in the focus ($\vec{f}$). There is also no temporal precedence relation between these elements. All elements are simultaneously accessible.

\end{itemize}

Based on these representations, each object in the domain of interest belongs to either one of these above vectors or {\em Others}. Each object in the above vectors consists of the following attributes:
\begin{itemize}
\item Semantic type of the object. For example, the semantic type could be a \textsf{House}  or a \textsf{Town}.
\item The attributes of the object. This is a domain dependent feature. A set of attributes is associated with each semantic type. For example, a house object has {\em Price, Size, Year Built}, etc. as its attributes. Furthermore, each object has visual properties that reflect the appearance of the object on the display such as Color of an object icon.
\item The identifier of the object. Each object has a unique name.
\item The selection probability. It refers to the probability that a given object is selected.  Depending on the interface design, a gesture could result in a list of potential referents. We use this selection probability to indicate the likelihood of an object selected by a gesture. The calculation of the selection probability is described later. For objects from the focus vector and the display vector, the selection probabilities are set to $1/N$ where $N$ is the total number of objects in the respective vector.
\item Temporal information. The relative temporal ordering information for the corresponding gesture. Instead of applying time stamps as in our previous work~\cite{chai:04a}, here we only use the index of gestures according to the order of their occurrences. If an object is selected by the first gesture, then its temporal information would be {\em 1}.
\end{itemize}

In addition to vectors that capture potential referents, for each user input, a vector that represents referring expressions from a speech utterance ($\vec{r}$) is also maintained. Each element (i.e., a referring expression) has the following information:
\begin{itemize}
\item The identifier of the potential referent indicated by the referring expression. For example, the identifier of the potential referent to the expression {\em house number eight} is a house object with an identifier {\em Eight}.
\item   The semantic type of the potential referents indicated by the expression. For example, the semantic type of the referring expression {\em this house} is {\em House}.
\item The number of potential referents as indicated by the referring expression or the utterance context. For example, a singular noun phrase refers to one object. A phrase like {\em three houses} provides the exact number of referents (i.e., 3).
\item Type dependent features. Any features associated with potential referents, such as {\em Color} and {\em Price}, are extracted from the referring expression.
\item The temporal ordering information indicating the order of referring expressions as they are uttered.  Again, instead of the specific time stamp, here we only use the temporal ordering information. If an utterance consists of $N$ consecutive referring expressions, then the temporal ordering information for each of them would be 1, 2, and up to $N$.
\item   The syntactic categories of the referring expressions. Currently, for each referring expression, we assign it to one of six syntactic categories (e.g., demonstrative and pronoun). Details are explained later.
\end{itemize}
These four vectors are updated after each user turn in the conversation based on the current user input and the system state (e.g., what is shown on the screen and what was identified as focus from the previous turn of the conversation).

\subsection{Algorithm}

\begin{figure}
\begin{center}
\epsfxsize=6in {\epsfbox{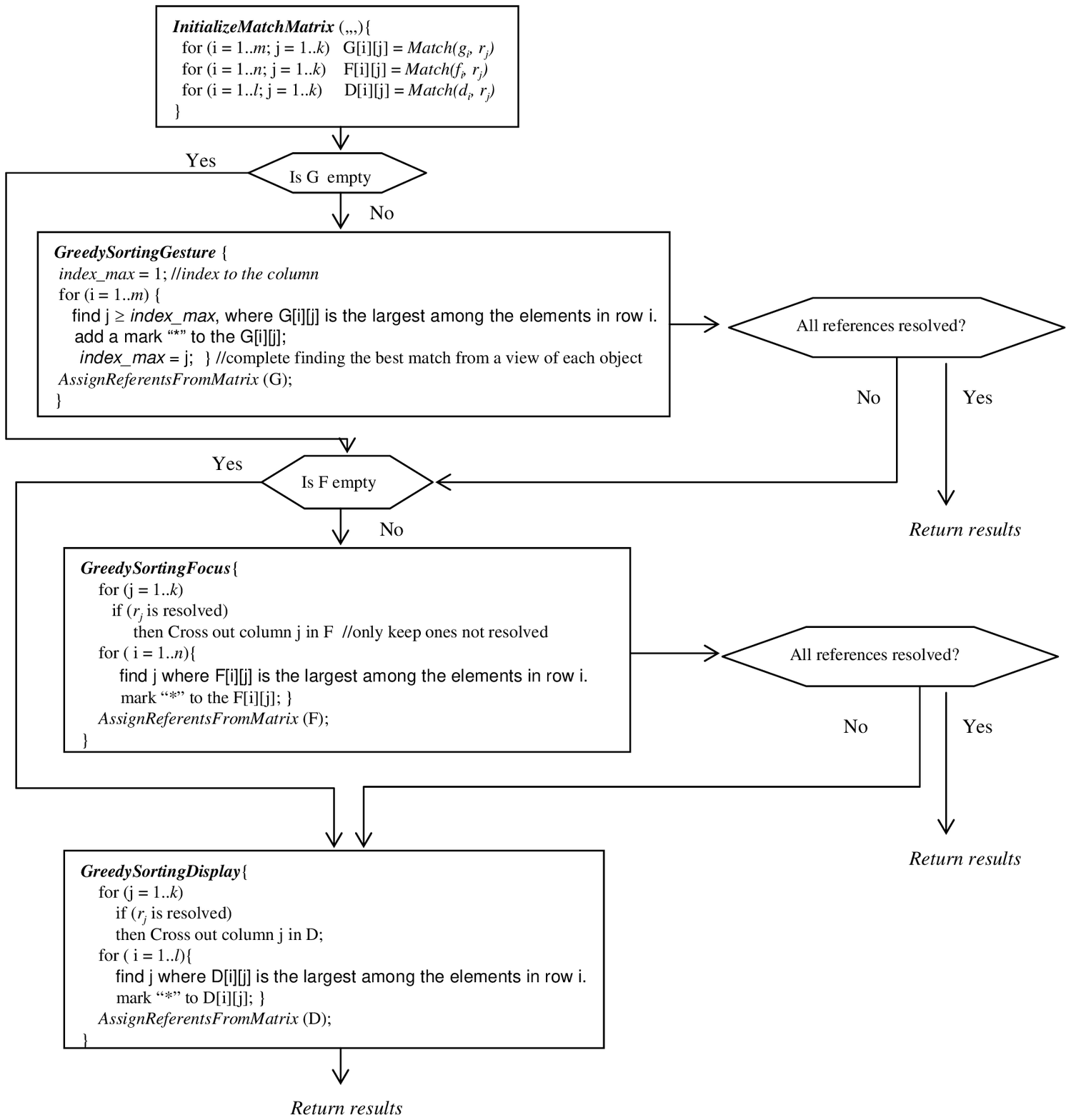}}
\epsfxsize=6in {\epsfbox{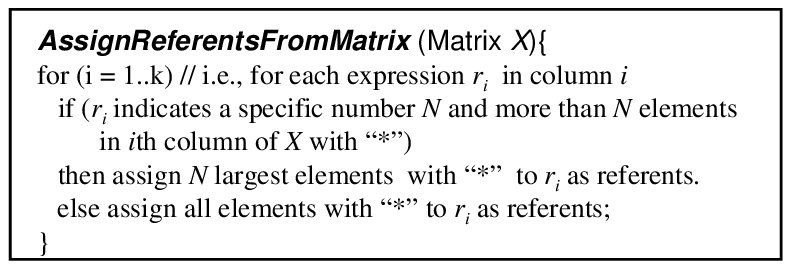}}
\caption{\small A greedy algorithm for multimodal reference resolution}
\label{fig:algorithm}
\end{center}
\end{figure}

The flow chart with the pseudo code of the algorithm is shown in Figure~\ref{fig:algorithm}. For each multimodal input at a particular turn in the conversation, this algorithm takes the inputs of a vector ($\vec{r}$) of referring expressions with size $k$, a gesture vector ($\vec{g}$ ) of size $m$, a focus vector of ($\vec{f}$ ) of size $n$, and a display vector ($\vec{d}$) of size $l$. It first creates three matrices $G[i][j]$, $F[i][j]$, and $D[i][j]$ to capture the scores of matching each referring expression from $\vec{r}$ to each object in the three vectors. Calculation of the matching score is described later. Note that, if any of the $\vec{g}$,$\vec{f}$, and $\vec{d}$ is empty, then the corresponding matrix (i.e., $G$, $F$, or $D$) is empty.  

Based on the matching scores in the three matrices, the algorithm applies a greedy search that is guided by our modified hierarchy as described earlier. Since {\em Gesture} has the highest status, the algorithm first searches the Gesture Matrix ($G$) that keeps track of matching scores between all referring expressions and all objects from gestures. It identifies the highest (or multiple highest) matching scores and assigns all possible objects from gestures to the expressions ({\em GreedySortingGesture}).

If more referring expressions are left to be resolved after gestures are processed, the algorithm looks at objects from the Focus Matrix ($F$) since {\em Focus} is the next highest cognitive status ({\em GreedySortingFocus}). If there are still more expressions to be resolved, then the algorithm looks at objects from the Display Matrix ($D$) ({\em GreedySortingDisplay}). Currently, our algorithm focuses on these three statuses. Certainly, if there are still more expressions to be resolved after all these steps, the algorithm can consult with proper name resolution. Once all the referring expressions are resolved, the system will output the results. For the next multimodal input, the system will generate four new vectors and then apply the greedy algorithm again.

Note that in {\em GreedySortingGesture}, we use {\em index-max} to keep track of the column index that corresponds to the largest matching value. As the algorithm incrementally processes each row in the matrix, this {\em index-max} should incrementally increase. This is because the referring expressions and the gesture should be aligned according to their order of occurrences. Since objects in the Focus Matrix and the Display Matrix do not have temporal precedence relations, {\em GreedySortingFocus} and {\em GreedySortingDisplay} do not use this constraint.

The reason we call this algorithm {\em greedy} is that it always finds the best assignment for a referring expression given a cognitive status in the hierarchy. In other words, this algorithm always makes the best choice for each referring expression one at a time according to the order of their occurrence in the utterance. One can imagine that a mistaken assignment made to an expression can affect the assignment of the following expressions.  Therefore, the greedy algorithm may not lead to a globally optimal solution. Nevertheless, the general user behavior following the guiding principles makes this greedy algorithm useful.

One major advantage of this greedy algorithm is that the use of the modified hierarchy can significantly prune the search space compared to the graph-matching approach. Given $m$ referring expressions and $n$ potential referents from various sources (e.g., gesture, conversation context, and visual display), this algorithm can find a solution in $O(mn)$. Furthermore, this algorithm goes beyond simple and precise inputs as illustrated by the decision list in Kehler (2000).  The scoring mechanism (described later) and the greedy sorting process accommodate both complex and ambiguous user inputs.

\subsection{Matching Functions}

An important component of the algorithm is the matching score between an object ($o$) and a referring expression ($e$). We use the following equation to calculate the matching score:
\begin{equation}
 Match(o,e) = [ \sum_{S \in \{G, F, D\}}P(o|S)*P(S|e)] * Compatibility(o,e)
 \end{equation}

In this formula, $S$ represents the possible associated status of an object $o$. It could have three potential values: $G$ (representing Gesture), $F$ (Focus), and $D$ (Display).
This function is determined by three components:
\begin{itemize}

\item The first, $P(o|S)$, is the object selectivity component that measures the probability of an object to be the referent given a status ($S$) of that object (i.e., gesture, focus, or visual display).

\item The second, $P(S|e)$, is the likelihood of status component that measures the likelihood of the status of the potential referent given a particular type of referring expression.

\item   The third, $Compatibility(o, e)$, is the compatibility component that measures the semantic and temporal compatibility between the object $o$ and the referring expression $e$.

\end{itemize}

Next we explain these three components in detail.

\subsubsection{Object Selectivity}

To calculate $P(o | S = Gesture)$, we use a function that takes into consideration of the distance between an object and the focus point of a gesture on the display~\cite{chai:04a}.

Given an object from {\em Focus} (i.e., not selected by any gesture), $P(o | S= Focus) = 1/N$, where $N$ is the total number of objects that are in the {\em Focus} vector. If an object is neither selected by a gesture, nor in the focus, but visible on the screen, then $P(o | S= Display) = 1/M$, where $M$ is the total number of objects that are in the {\em Display} vector. Currently, we only applied the simplest uniform distribution for objects in focus and on the graphical display. In the future, we intend to incorporate the recency in conversation discourse to model $P(o | S= Focus)$ and use visual prominence (e.g., based on visual characteristics) to model $P(o | S= Display)$. Note that, as discussed earlier in Section~\ref{sec:representation}, each object is associated with only one of the three statuses. In other words, for a given object $o$, only one of $P(o|S=Gesture)$, $P(o|S=Focus)$, and $P(o|S=Display)$ is non-zero.

\subsubsection{Likelihood of Status}
Motivated by the Givenness Hierarchy and earlier work~\cite{kehler:00} that the form of referring expressions can reflect the cognitive status of referred entities in a user's mental model, we use the likelihood of status to measure the probability of a reflected status given a particular type of referring expression.
In particular, we use the data reported in Kehler (2000) to derive the likelihood of the status of potential referents given a particular type of referring expression $P(S|e)$. We categorize referring expressions into the following six categories:
\begin{itemize}
\item Empty: no referring expression is used in the utterance.
\item Pronouns: such as {\em it}, {\em they}, and {\em them}
\item   Locative adverbs: such as {\em here}  and {\em there}
\item   Demonstratives: such as {\em this}, {\em that}, {\em these}, and {\em those}
\item   Definite Noun Phrases: noun phrases with the definite article {\em the}
\item Full noun phrases: other types such as proper nouns.
\end{itemize}

Table~\ref{tab:likelihood} shows the estimated $P(S|e)$. Note that, in the original data provided by Kehler (2000), there is zero count for a certain combination of a referring type and a referent status. These zero counts result in zero probability in the table. We did not use any smoothing techniques to re-distribute the probability mass. Furthermore, there is no probability mass assigned to the status {\em Others}. 

\begin{table}
\center
\begin{tabular}{|c|l|l|l|l|l|l|} \hline
    $P(S|E) $    & {\bf Empty} & {\bf Pronoun}   & {\bf Locative} & {\bf Demonstratives} & {\bf Definite} & {\bf Full} \\ \hline \hline
{\bf Visible }  &  0        &  0       &  0      &  0        &  0       &  0  \\ \hline
{\bf Focus}     & 0.56      & 0.85     &0.57     &0.33       &0.07      & 0.47 \\ \hline
{\bf Gesture }  & 0.44      & 0.15     &0.43     &0.67       &0.67      & 0.16 \\ \hline
{\bf Sum}               & 1     & 1    &1     & 1     & 1    &1  \\ \hline \hline
\end{tabular}
\caption{Likelihood of status of referents given a particular type of expression}
\label{tab:likelihood}
\end{table}

\subsubsection{Compatibility Measurement}
\label{sec:compatibility}

The term $Compatibility(o, e)$ measures the compatibility between  an object $o$ and a referring expression $e$. Similar to the compatibility measurement in our earlier work~\cite{chai:04c}, it is defined by a multiplication of many factors in the following equation:
\begin{equation}
Compatibility (o, e) =  Id(o, e) *Sem(o, e) * \prod_k Attr_k(o, e)*Temp(o, e)
\end{equation}
In this equation:
\begin{description}
\item [$Id(o,e)$] It captures the compatibility between  the identifier (or name) for $o$ and the identifier (or name) specified in $e$. It indicates that the identifier of the potential referent, as expressed in a referring expression, should match the identifier of the true referent. This is particularly useful for resolving proper nouns. For example, if the referring expression is house number eight, then the correct referent should have the identifier number eight. $Id(o, e) = 0$ if the identities of $o$ and $e$ are different. $Id(o, e) = 1$ if the identities of $o$ and $e$ are either the same or one/both of them unknown.

\item [$Sem (o, e)$] It captures the semantic type compatibility between $o$ and $e$.  It indicates that the semantic type of a potential referent as expressed in the referring expression should match the semantic type of the correct referent. $Sem (o, e) = 0$ if the semantic types of $o$ and $e$ are different. $Sem (o, e) = 1$ if they are the same or unknown.

\item [$Attr_k(o, e)$] It captures the type-specific constraint concerning a particular semantic feature (indicated by the subscript $k$). This constraint indicates that the expected features of a potential referent as expressed in a referring expression should be compatible with features associated with the true referent. For example, in the referring expression {\em the Victorian house}, the {\em style} feature is {\em Victorian}.  Therefore, an object can only be a possible referent if the style of that object is {\em Victorian}.  Thus, we define the following: $Attr_k(o, e) = 0$ if both $o$ and $e$ have the feature $k$ and the values of the feature $k$ are not equal. Otherwise, $Attr_k(o, e) = 1$.

\item [$Temp(o, e)$] It captures the temporal compatibility between $o$ and $e$.  Here we only consider the temporal ordering between speech and gesture.  Specifically, the temporal compatibility is defined as the following:
\begin{equation}
 Temp(o,e) = exp(-|OrderIndex(o) - OrderIndex(e)|)
 \end{equation}
The order when the speech and the accompanying gestures occur is important in deciding which gestures should be aligned with which referring expressions. The order in which the accompanying gestures are introduced into the discourse should be consistent with the order in which the corresponding referring expressions are uttered. For example, suppose a user input consists of three gestures $g_1, g_2, g_3$ and two referring expressions, $s_1, s_2$. It will not be possible for $g_3$ to align with $s_1$ and $g_2$ to align with $s_2$. Note that, if the status of an object is either  {\em Focus} or {\em Visible}, then $Temp(o, e) = 1$.  This definition of temporal compatibility is different from the function used in our previous work~\cite{chai:04c} that takes real time stamps into consideration.
Section~\ref{sec:eval-alignment} shows different performance results based on different temporal compatibility functions.
\end{description}

\subsection{An Example}

\begin{figure}
\begin{center}
\epsfxsize=3.5in {\epsfbox{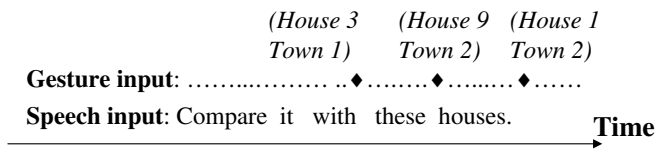}}
\caption{\small An example of a complex input}
\label{fig:example}
\end{center}
\end{figure}

\begin{figure}
\begin{center}
\epsfxsize=4.5in {\epsfbox{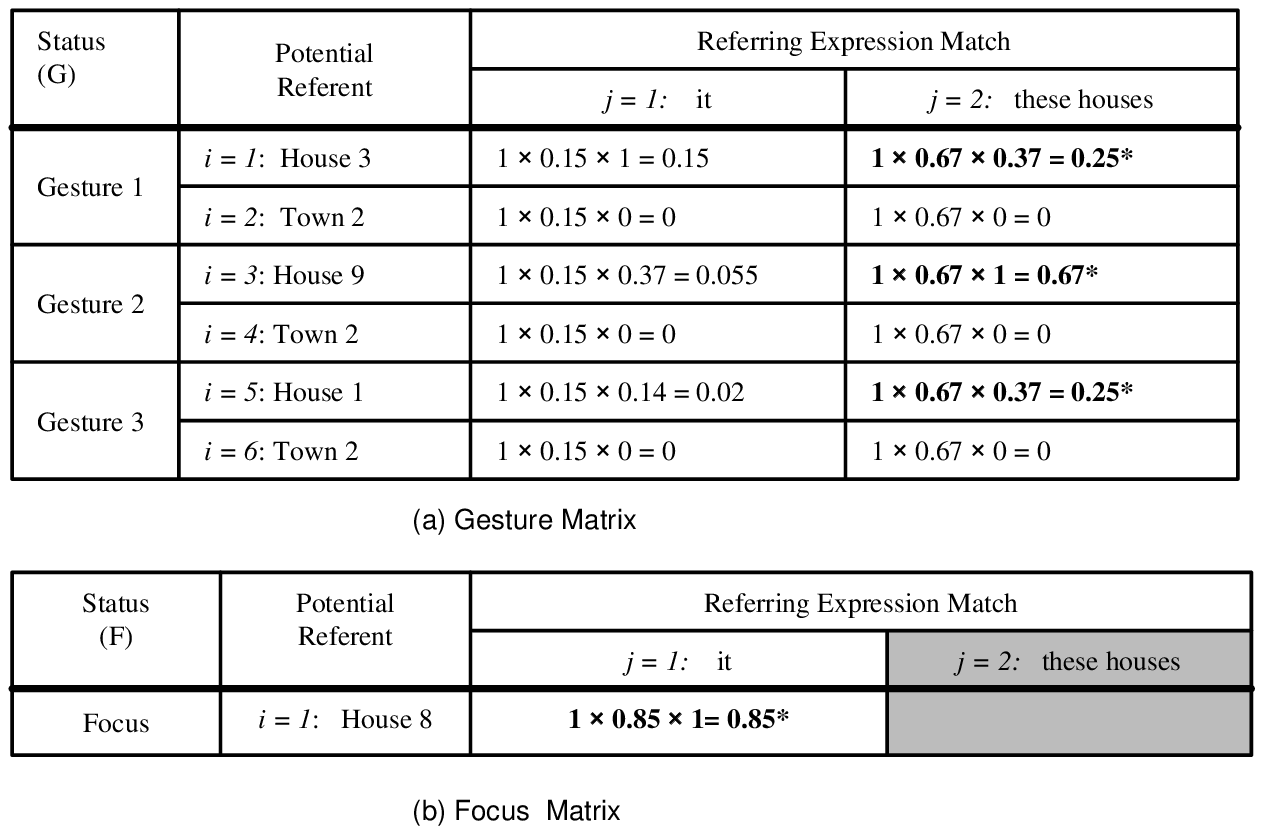}}
\caption{\small The Gesture Matrix (a) and Focus Matrix (b) for processing the example in Figure~\ref{fig:example}. Each cell in the {\em Referring Expression Match} columns corresponds to an instantiation of the matching function.}
\label{fig:exampleTable}
\end{center}
\end{figure}

Figure~\ref{fig:example} shows an example of a complex input that involves multiple referring expressions and multiple gestures. Because the interface displays house icons on top of town icons, a point (or circle) could result in both a house and a town object. In this example, the first gesture results in both {\em House 3} and {\em Town 1}.  The second gesture results in {\em House 9} and {\em Town 2}, and the third results in {\em House 1} and {\em Town 2.} Suppose before this input takes place, {\em House 8} is highlighted on the screen from the previous turn of conversation (i.e., {\em House 8} is in the focus). Furthermore, there are eight other objects visible on the screen.
To resolve referents to the expressions {\em it} and {\em these houses}, the greedy algorithm takes the following steps:
\begin{enumerate}
\item The four input vectors,  $\vec{g}$,$\vec{f}$, $\vec{d}$, and $\vec{r}$  are created with lengths 6, 1, 8, 2, respectively to represent six objects in the gesture vector, one object in the focus, eight more objects on the graphical display, and two referring expressions used in the utterance.
\item   Gesture Matrix $G_{62}$, Focus Matrix $F_{12}$, and Display Matrix $D_{82}$ are created.
\item These three matrixes are then initialized by Equation 2.  Figure~\ref{fig:exampleTable} shows the resulting Gesture Matrix. The probability values of $P(S|e)$ come from Table~\ref{tab:likelihood}. The difference in the compatibility values for the house objects in the Gesture Matrix is mainly due to the temporal ordering compatibilities.
\item   Next the {\em GreedySortingGesture} procedure is executed. For each row in Gesture Matrix, the algorithm finds the largest legitimate value and marks the corresponding cell with *. The {\em legitimate} means that the corresponding cell for the row $i+1$ has to be either on the same column or the column to the right of the corresponding cell in row i. These values are shown in bold in Figure~\ref{fig:exampleTable}(a). Next, starting from each column, the algorithm checks for each referring expression whether any $*$ exists in its corresponding column. If so, those objects with $*$ are assigned to the referring expressions based on the number constraints. In this case, since no specific number is given in the referring expression {\em these houses}, all three marked objects are assigned to {\em these houses}.
\item   After {\em these houses}, there is still {\em it} left to be resolved.  Now the algorithm continues to execute {\em GreedySortingFocus}. The Focus Matrix prior to executing {\em GreedySortingFocus} is shown in Figure~\ref{fig:exampleTable}(b). Note that since {\em these houses} is no longer considered, its corresponding column is deleted from the Focus Matrix. Similar to the previous step, the largest non-zero match value is marked (shown in bold in Figure~\ref{fig:exampleTable}(b)) and assigned to the remaining referring expression {\em it}.
\item   The resulting Display Matrix is not shown because at this point, all referring expressions are resolved.
\end{enumerate}

\section{Evaluation}
We use the data collected from our previous work \cite{chai:04c} to evaluate this greedy algorithm. The questions addressed in our evaluation are the following:
\begin{itemize}
\item What is the impact of temporal alignment between speech and gesture on the performance of the greedy algorithm?
\item What is the role of modeling the cognitive status in the greedy algorithm?
\item How effective is the greedy algorithm compared to the graph matching algorithm (Section~\ref{sec:graphmatching})?
\item   What error sources contribute to the failure in real-time reference resolution?
\item How is the greedy algorithm compared to the finite state approach (Section~\ref{sec:fst}) and the decision list approach (Section~\ref{sec:dlist})?
\end{itemize}

\begin{table}
\center
\begin{tabular}{|l|l|l|l|l|l|l|l|} \hline
                                    & {\bf $g_1$} & {\bf $g_2$}   & {\bf $g_3$} & {\bf $g_4$} & {\bf $g_5$} & {\bf $g_6$}  & Total \\
                                    & no          & one           & mult.    & one         & mult.    & pts \&   &     Num            \\
                                    & gest.       & pt            & pts       & cir         & cirs     & cirs      &                 \\ \hline \hline
{\bf $s_1$}: $the (adj)^*(N|Ns)$     &  2   & 8 & 0 & 2 & 0 & 1 & 13         \\ \hline
{\bf $s_2$}: $(this|that)(adj^*) N$   &   4 & 43 & 3 & 33 & 1 & 7 & 91       \\ \hline
{\bf $s_3$}: $(these|those)(num^+)(adj^*)Ns$     &    0 & 0 & 0 & 31 & 0 & 5 & 36      \\ \hline
{\bf $s_4$}: $it|this|that|(this|that|the)adj^*one$   & 3 & 8 & 0 & 10 & 0 & 0 & 21         \\ \hline
{\bf $s_5$}: $(these|those)num^+adj^*ones|them$    &    0 & 0 & 0 & 2 & 0 & 0 & 2       \\ \hline
{\bf $s_6$}: $here|there$  &        1 & 1 & 0 & 5 & 0 & 0 & 7       \\ \hline
{\bf $s_7$}: $empty\ expression$    &     1 & 1 & 0 & 1 & 0 & 0 & 3        \\ \hline
{\bf $s_8$}: $proper\ nouns$                &     1 & 5 & 3 & 3 & 0 & 3 & 15        \\ \hline
{\bf $s_9$}: $multiple\ expressions$                &     1 & 0 & 4 & 11 & 13 & 2 & 31         \\ \hline
{\bf Total Num}:                &        13 & 66 & 10 & 98 & 14 & 18 & 219       \\ \hline \hline
\end{tabular}
\caption{Detailed description of user referring behavior}
\label{tab:det-behavior}
\end{table}

\subsection{Experiment Setup}
The evaluation data were collected from eleven subjects who participated in our study. 
Each of the subjects was asked to interact with the system using both speech and gestures (e.g., pointing and circle) to accomplish five tasks related to real estate information seeking. The first task was to find the least expensive house in the most populated town. In order to accomplish this task, the user would have to first find the town that has the highest population and then find the least expensive house in this town. The next task involved obtaining a description of the house located in the previous task. The next task was to compare the house that was located in the first task with all of the houses in a particular town in terms of price. Additionally, the least expensive house in this second town should be determined.  Another task was to find the most expensive house in a particular town. The last task involved comparing the resulting houses of the previous four tasks. For this last task, the previous four tasks may have to be completely or partially repeated. These tasks were designed so that users were required to explore the interface to acquire various types of information.

The acoustic model for each subject was trained individually to minimize speech recognition errors. The study session was videotaped to capture both audio and video on the screen movement (including gestures and system responses). The IBM Viavoice speech recognizer was used to process each speech input.

Table~\ref{tab:det-behavior} provides a detailed description of the referring behavior observed in the study. The columns indicate whether no gesture, one gesture (pointing or circle), or multiple gestures are involved in a multimodal input. The rows indicate the type of referring expressions in a speech utterance. Each table entry shows the number of a particular combination of speech and gesture inputs.

Table~\ref{tab:sum-behavior} summarizes Table~\ref{tab:det-behavior} in terms of whether  no gesture, one gesture, or multiple gestures (shown as columns) and whether  no referring expression, one referring expression, or multiple referring expressions (shown as rows) are involved in the input.  Note that in this table an intended input is counted as one input even if this input may be split into a few turns by our system during the run time.

\begin{table}
\center
\begin{tabular}{|l|l|l|l|l|} \hline
                                    & {\bf $G_0$: No} & {\bf $G_1$: One}   & {\bf $G_2$: Multi-}   & Total \\
                                    &  {\bf Gesture }     & {\bf Gesture }          & {\bf Gesture }      &     Num            \\ \hline \hline
{\bf $S_0$}: No referring expression     &     1   $_{(a)}$ & 2  $_{(a)}$ & 0 $_{(c)}$& 3       \\ \hline
{\bf $S_1$}: One referring expression  &      11  $_{(a)}$ & 151 $_{(b)}$& 23 $_{(c)}$& 185          \\ \hline
{\bf $S_2$}: Multiple referring expressions   &   1 $_{(c)}$& 11  $_{(c)}$ & 19  $_{(c)}$ & 31                \\ \hline
{\bf Total Num}:            &    13 & 164 & 42 & 219              \\ \hline
\end{tabular}
\caption{Summary of user referring behavior}
\label{tab:sum-behavior}
\end{table}

Based on Table~\ref{tab:sum-behavior}, we further categorize user inputs into the following three categories:
\begin{itemize}
\item Simple Inputs with One-Zero Alignment: inputs that contain no speech referring expression with no gesture (i.e.,$<S_0, G_0>$), one referring expression with zero gesture (i.e.,$<S_1, G_0>$), and no referring expression with one gesture (i.e.,  $<S_0, G_1>$). These types of inputs require the conversation context or visual context to resolve references. One example of this type is the $U_2$ in Table~\ref{tab:fragment}. From our data, a total of 14 inputs belong to this category (marked $(a)$ in Table~\ref{tab:sum-behavior}). 

\item Simple Inputs with One-One Alignment: inputs that contain exactly one referring expression and one gesture (i.e., $<S_1, G_1>$). These types of inputs can be resolved mostly by combining gesture and speech using multimodal fusion.  A total of 151 inputs belong to this category (marked $(b)$ in Table~\ref{tab:sum-behavior}). 

\item Complex Inputs: inputs that contain more than one referring expression and/or gesture. This corresponds to the entry $<S_1, G_2>$, $<S_2, G_0>$,$<S_2, G_1>$, and $<S_2, G_2>$ in Table~\ref{tab:sum-behavior}. One example of this type is $U_3$ in Table~\ref{tab:fragment}. A total of 54 inputs belong to this category (marked $(c)$ in Table~\ref{tab:sum-behavior}). These types of inputs are particularly challenging to resolve.
\end{itemize}

In this section, we will focus on different performance evaluations based on these three types of referring behaviors.

\subsection{Temporal Alignment Between Speech and Gesture}
\label{sec:eval-alignment}

In multimodal interpretation, how to align speech and gesture based on their temporal information is an important question. This is especially the case for complex inputs where  a multimodal input consists of multiple referring expressions and multiple gestures. We evaluated different temporal compatibility functions for the greedy approach. In particular, we compared the following three functions:
\begin{itemize}
\item The {\em ordering} temporal constraint as in Equation 4.
\item The {\em absolute} temporal constraint as defined by the following formula:
\begin{equation}
 Temp(o,e) = exp(-|BeginTime(o)-BeginTime(e)|)
\end{equation}
    Here, the absolute timestamps of the potential referents (e.g., indicated by a gesture) and the referring expressions are used instead of the relative orders of relevant entities in a user input.
\item The {\em combined} temporal constraint that combines the two aforementioned constraints, giving each equal weight in determining the compatibility score between an object and a referring expression.
\end{itemize}

\begin{table}
\center
\begin{tabular}{|l|l|l|l|} \hline
     No. Correctly Resolved      & {\em Ordering}  & {\em Absolute}  & {\em Combined} \\ \hline \hline
 Simple One-Zero Alignment       &        5        &       5          &     5          \\ \hline
 Simple One-One Alignment       &       104         &   104              &      104         \\ \hline
 Complex           &        24        &       19          &       23        \\ \hline
 Total                 &   133             &  128               &     132          \\ \hline
  Accuracy             &        60.7\%        &    58.4\%             &    60.3\%           \\ \hline \hline
\end{tabular}
\caption{Performance comparison based on different temporal compatibility functions}
\label{tab:temporal-comparison}
\end{table}

The results are shown in Table~\ref{tab:temporal-comparison}. Different temporal constraints only affect the processing of complex inputs. The ordering temporal constraint worked slightly better than the absolute temporal constraint. 
In fact, temporal alignment between speech and gesture is often one of the problems that may affect interpretation results. Previous studies have found the gestures tend to occur before the corresponding speech unit takes place \cite{oviatt:97}. The findings suggest that users tend to tap on the screen first and then start the speech utterance. This behavior was observed in a simple command based system \cite{oviatt:97} where each speech unit corresponds with a single gesture (i.e., the simple inputs in our work).

From our study, we found that temporal alignment between gesture and corresponding speech units is still an issue that needs to be further investigated in order to improve the robustness in multimodal interpretation. Table~\ref{tab:temp-behavior} shows the percentage of different temporal relations observed in our study. The rows indicate whether there is an overlap between speech referring expressions and their accompanied gestures. The columns indicate whether the speech (more precisely, the referring expressions) or the gesture occurred first. Consistent with the previous findings \cite{oviatt:97}, in most cases (85\% of time), gestures occurred before the referring expressions were uttered. However, in 15\% of the cases the speech referring expressions were uttered before the corresponding gesture occurred. Among those cases, 8\% had an overlap between the referring expressions and the gesture and 7\% had no overlap.

\begin{table}
\center
\begin{tabular}{|l|l|l|l|} \hline
                & Speech First      & Gesture First      &  Total            \\ \hline \hline
Non-overlap    &     7\%              &  45\%             &  52\%                   \\ \hline
Overlap        &      8\%             &  40\%             &  48\%           \\ \hline
Total :             &      15\%             &   85\%            &   100\%                \\ \hline\hline
\end{tabular}
\caption{Overall temporal relations between speech and gesture}
\label{tab:temp-behavior}
\end{table}

\begin{figure}
\begin{center}
\epsfxsize=4.5in {\epsfbox{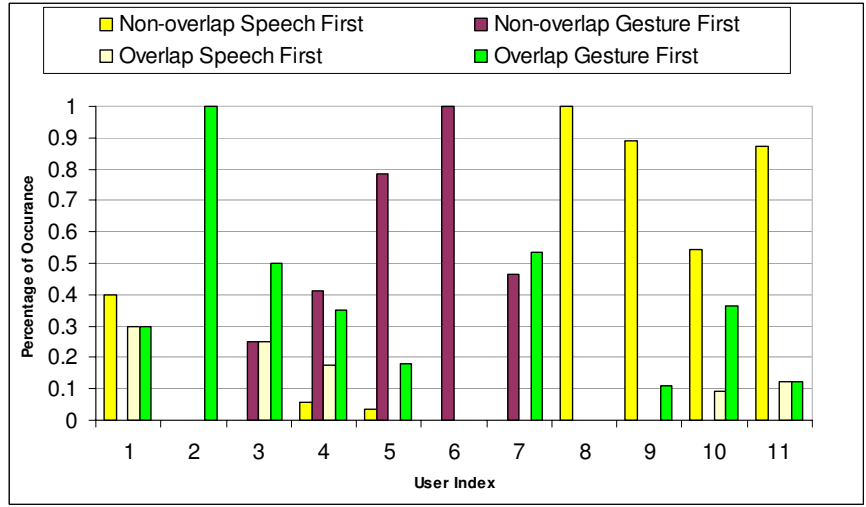}}
\caption{\small Temporal alignment behavior from our user study}
\label{fig:alignment}
\end{center}
\end{figure}

Furthermore, although multimodal behaviors such as sequential (i.e., non-overlap) or simultaneous (e.g., overlap) integration are quite consistent during the course of interaction~\cite{oviatt:03}, there are some exceptions. Figure~\ref{fig:alignment}  shows the temporal alignments from individual users in our study. User 2 , User 6, and User 8 maintained a consistent behavior in that User 2's gesture always happened before and overlapped with the corresponding speech referring expressions; User 6's gesture always occurred ahead of the speech expressions without overlapping; and User 8's speech referring expressions always occurred before the corresponding gestures (without any overlap).  The other users exhibited varied temporal alignment between speech and gesture during the interaction.  It will be difficult for a system using pre-defined temporal constraints to anticipate and accommodate all these different behaviors.  Therefore, it is desirable to have a mechanism that can automatically learn the user behavior of alignment and automatically adjust to that behavior. 

One potential approach is to introduce a calibration process before real human computer interaction. In this calibration process, two tasks will be performed by a user. In the first task,  the user will be asked to describe objects on the graph display with both speech and deictic gestures. In the second task, the user will be asked to respond to the system questions by using both speech and deictic gestures. The reason to have users perform these two tasks is to identify whether there is any difference between user initiated inputs and system initiated user responses. Based on these tasks, the temporal relations between the speech units and corresponding gestures can be captured and used in the real-time interaction.  

\subsection{The Role of Cognitive Principles}
To further examine the role of modeling cognitive status in multimodal reference, we compared the two configurations of the greedy algorithm. The first configuration is based on the matching score defined in Equation 2, which incorporates the cognitive principles described earlier.  The second configuration only uses the matching score that is completely dependent on the compatibility between a referring expression and a gesture (i.e., Section~\ref{sec:compatibility}) without using the cognitive principles (i.e., $P(o|S)$ and $P(S|e)$ are not included in Equation 2). 

Table~\ref{tab:results-cog} shows the comparison results in terms of these two configurations. The algorithm using the cognitive principles outperforms the algorithm that does not use the cognitive principles by more than 15\%. The performance difference applies to both simple inputs with one-one alignment and complex inputs. The results indicate that modeling cognitive status can potentially improve reference resolution performance.

\begin{table}
\center
\begin{tabular}{|l|c|c|} \hline
 No. Correctly Resolved               & with Cognitive Principles      & without Cognitive Principles                 \\ \hline \hline
 Simple One-Zero Alignment   &     5             &        5                       \\ \hline               
Simple One-One Alignment   &    104              &         92                       \\ \hline
Complex   &      24             &          18              \\ \hline
Total            &     133             &        115               \\  \hline\hline      

\end{tabular}
\caption{The role of cognitive principles in the greedy algorithm}
\label{tab:results-cog}
\end{table}

\subsection{Greedy Algorithm versus Graph-matching Algorithm}
We further compared the greedy algorithm and the graph-matching algorithm in terms of performance and runtime. 
Table~\ref{tab:performance-comparison} shows the performance comparison. Overall, the greedy algorithm performs comparably with the graph-matching algorithm.

\begin{table}
\center
\begin{tabular}{|l|l|l|l|l|l|} \hline
        & Total Num   &
            \multicolumn{2}{c|}{Graph-matching} &
            \multicolumn{2}{c|}{Greedy} \\ \cline{3-6}
               &      &  Num & \% & Num & \%  \\ \hline \hline
     Total                    &  219   &  130 &  59.4\%  & 133   & 60.7\%        \\ \hline\hline
    Simple One-Zero Alignment   &  14      &  7    &   50.0\%       &  5      & 35.7\%           \\ \hline
    Simple One-One Alignment   &  151      & 104     & 68.9\%         & 104      &    68.9\%           \\ \hline 
              Complex           &  54      &  19    &  35.2\%        &    24   &   44.4\%         \\ \hline\hline
\end{tabular}
\caption{Performance comparison between the graph-matching algorithm and the greedy algorithm}
\label{tab:performance-comparison}
\end{table}

To compare the runtime, we ran each algorithm on each user 10 times where each input was run 100 times. In other words, each user input was run 1000 times by each algorithm to get the average runtime measurement. This experiment was done on a UltraSPARC-III server with 750MHz and 64bit. 

Both the greedy algorithm and the graph-matching algorithm have the same function calls to process speech inputs (e.g., parsing) and gesture inputs (e.g., identify potentially intended objects). The difference between these algorithms are the specific implementations regarding graph creation and matching as in the graph-matching algorithm and the greedy search as in the greedy algorithm. As a result, the average time for the greedy algorithm to process simple inputs and complex inputs are 17.3 milliseconds and 21.2 milliseconds respectively. The average time for the graph matching algorithm to process simple and complex inputs are 22.3 milliseconds and 24.8 milliseconds respectively. These results show that on average the greedy algorithm runs slightly faster than the graph-matching algorithm given our current implementation, although in the worst case, the graph-matching algorithm is asymptotically more complex.

\subsection{Real-time Error Analysis}
\label{sec:erroranalysis}

To understand the bottleneck in real-time multimodal reference resolution, we examined the error cases where the algorithm failed to provide correct referents. 

Like in most spoken dialog systems, speech recognition is a major bottleneck. Although we have trained each user's acoustic model individually, the speech recognition rate is still very low. Only 127 of inputs had correctly recognized referring expressions. Among these inputs, 103 of them were resolved with correct referents.  Fusing inputs from multiple modalities together can sometimes compensate for the recognition errors \cite{oviatt:96}. Among 92 inputs in which referring expressions were incorrectly recognized, 29 of them were correctly assigned referents due to the mutual disambiguation. A mechanism to reduce the recognition errors, especially by utilizing information from other modalities, will be important to provide a robust solution for real time multimodal reference resolution.

The second source of errors comes from another common problem in most spoken dialog systems, namely out-of-vocabulary words. For example, {\em area} was not in our vocabulary. So the additional semantic constraint expressed by {\em area} was not captured. Therefore, the system could not identify whether a house or a town was referred to when the user uttered {\em this area}. It is important for the system to have a capability to acquire knowledge (e.g., vocabulary) dynamically by utilizing information from other modalities and the interaction context. Furthermore, the errors also came from a lack of understanding of spatial relations (as in {\em the house just close to the red one}) and superlatives (as in {\em the most expensive house}). Algorithms for aligning visual features to resolve spatial references are desirable~\cite{gorniak:04}.

In addition to these two main sources, some errors are caused by unsynchronized inputs. Currently, we use an idle status (i.e., 2 seconds with no input from either speech or gesture) as the boundary to delimit an interaction turn. Two types of out of synchronization were observed. The first type is unsynchronized inputs from the user (such as a big pause between speech and gesture) and the other comes from the underlying system implementation.  The system captures speech inputs and gesture inputs from two different servers through a TCP/IP protocol. A communication delay sometimes split one synchronized input into two separate turns of inputs (e.g., one turn was speech input alone and the other turn was gesture input alone). A better engineering mechanism for synchronizing inputs is desired.

The disfluencies from the users also accounted for a small number of errors. The current algorithm is incapable of distinguishing disfluent cases from normal cases. Fortunately, the disfluent situations did not occur frequently in our study (only 6 inputs with disfluency). This is consistent with the previous findings that speech disfluency rate is lower in human machine conversation than in spontaneous speech~\cite{Brennan:2000}. During human-computer conversation, users tend to speak carefully and utterances tend to be short. Recent findings indicated that gesture patterns could be used as an additional source to identify different types of speech disfluencies during human-human conversation \cite{chen:02}. Based on our limited cases, we found that gesture patterns could be indicators of speech disfluencies when they did occur. For example, if a user says {\em show me the red house (point to house A), the green house (still point to the house A)}, then the behavior of pointing to the same house with different speech description usually indicates a repair. Furthermore, gestures also involve disfluencies; for example, repeatedly pointing to an object is a gesture repetition. Failure in identifying these disfluencies caused problems with reference resolution. It will be ideal to have a mechanism that can identify these disfluencies using multimodal information.

\subsection{Comparative Evaluation with Two Other Approaches}
To further examine how the greedy algorithm is compared to the finite state approach (Section~\ref{sec:fst}) and the decision list approach (Section~\ref{sec:dlist}), we conducted a comparative evaluation. In the original finite state approach, the N-best speech hypotheses are maintained in the speech tape.  In our data here, we only had the best speech hypothesis for each speech input. Therefore, we manually updated some incorrectly recognized words so that the finite state approach would not be penalized because of the lack of N-best speech hypotheses~\footnote{Note that we only corrected those inputs where there was a direct correspondence between the recognized words and transcribed words to maintain the consistency of timestamps.}. The modified data were used in all three approaches. Table~\ref{tab:approach-comparison} shows the comparison results.

\begin{table}
\center
\begin{tabular}{|c|c|c|c|} \hline
 No. Correctly Resolved        &  {\bf Greedy}     &  {\bf Finite State}     &   {\bf Decision List} \\ \hline \hline
Simple Inputs with one-one alighment         &   116   & 115  & 88             \\ \hline
Simple Inputs with zero-one alighment        &  8   & 0  & 12             \\ \hline
Complex Inputs               & 24     & 13  & 0        \\ \hline
Total  & 148  & 128  & 100 \\ \hline\hline
\end{tabular}
\caption{Performance comparison with two other approaches}
\label{tab:approach-comparison}
\end{table}

As shown in this table, the greedy algorithm correctly resolved more inputs than the finite state approach and the decision list approach. 
The major problem with the finite state approach is that it does not incorporate conversation context in the finite state transducer. This problem contributes to the failure in resolving simple inputs with zero-one alignment and some of the complex inputs. The major problem with the decision list approach, as described earlier, is the lack of capabilities to process ambiguous gestures and complex inputs.

Note that the greedy algorithm is not an algorithm to obtain the full semantic interpretation of a multimodal input.  But rather it is an algorithm specifically for reference resolution, which uses information from context and gesture to resolve speech referring expressions. In this regard, the greedy algorithm is different from the finite state approach whose goal is to get a full interpretation of user inputs and reference resolution is only a part of this process.

\section{Conclusion}

Motivated by earlier investigation on the cognitive status in human machine interaction,  this paper describes a greedy algorithm that incorporates the cognitive principles underlying human referring behavior to resolve a variety of references during human machine multimodal interaction. 
 In particular, this algorithm relies on the theories of Conversation Implicature and Givenness Hierarchy to effectively guide the system in searching for potential referents. 
Our empirical studies have shown that modeling the  form of referring experssions and its implication on the cognitive status can achieve better results than the algorithm that only considers the compatibility between referring expressions and potential referents.  
This greedy algorithm can efficiently achieve comparable performance as a previous optimization approach based on graph-matching.
Furthermore, because this greedy algorithm handles a variety  of user inputs ranging from precise to ambiguous and from simple to complex, it outperforms the finite state approach and the decision list approach in our experiments. 
Because of its simplicity and generality, this approach has a potential to improve the robustness of multimodal interpretation. We have learned from this investigation that prior knowledge from linguistic and cognitive studies can be very beneficial in designing efficient and practical algorithms for enabling multimodal human machine communication.

\acks{This work was supported by a NSF CAREER award IIS-0347548. The authors would like to thank anonymous reviewers for their valuable comments and suggestions.}



\vskip 0.2in
\bibliography{references}

\begin{thebibliography}{}

\bibitem[\protect\BCAY{Bolt}{Bolt}{1980}]{bolt:80}
Bolt, R. \BBOP1980\BBCP.
\newblock \BBOQ Put that there: Voice and gesture at the graphics
  interface\BBCQ\
\newblock {\Bem Computer Graphics}, {\Bem 14\/}(3), 262--270.

\bibitem[\protect\BCAY{Brennan}{Brennan}{2000}]{Brennan:2000}
Brennan, S. \BBOP2000\BBCP.
\newblock \BBOQ Processes that shape conversation and their implications for
  computational linguistics\BBCQ\
\newblock In {\Bem Proceedings of 38th Annual Meeting of ACL}, \BPGS\ 1--8.

\bibitem[\protect\BCAY{Byron}{Byron}{2002}]{byron:02}
Byron, D. \BBOP2002\BBCP.
\newblock \BBOQ Resolving pronominal reference to abstract entities\BBCQ\
\newblock In {\Bem Proceedings of 40th Annual Meeting of ACL}, \BPGS\ 80--87.

\bibitem[\protect\BCAY{Cassell, Bickmore, Billinghurst, Campbell, Chang,
  Vilhjalmsson,\ \BBA\ Yan}{Cassell et~al.}{1999}]{cassell:90}
Cassell, J., Bickmore, T., Billinghurst, M., Campbell, L., Chang, K.,
  Vilhjalmsson, H., \BBA\ Yan, H. \BBOP1999\BBCP.
\newblock \BBOQ Embodiment in conversational interfaces: Rea\BBCQ\
\newblock In {\Bem Proceedings of the CHI'99}, \BPGS\ 520--527.

\bibitem[\protect\BCAY{Chai, Hong, Zhou,\ \BBA\ Prasov}{Chai
  et~al.}{2004}]{chai:04c}
Chai, J., Hong, P., Zhou, M., \BBA\ Prasov, Z. \BBOP2004\BBCP.
\newblock \BBOQ Optimization in multimodal interpretation\BBCQ\
\newblock In {\Bem Proceedings of 42nd Annual Meeting of Association for
  Computational Linguistics (ACL)}, \BPGS\ 1--8.

\bibitem[\protect\BCAY{Chai, Prasov, Blaim,\ \BBA\ Jin}{Chai
  et~al.}{2005}]{chai:05}
Chai, J., Prasov, Z., Blaim, J., \BBA\ Jin, R. \BBOP2005\BBCP.
\newblock \BBOQ Linguistic theories in efficient multimodal reference
  resolution: An empirical study\BBCQ\
\newblock In {\Bem Proceedings of The 10th International Conference on
  Intelligent User Interfaces(IUI)}, \BPGS\ 43--50.

\bibitem[\protect\BCAY{Chai, Prasov,\ \BBA\ Hong}{Chai
  et~al.}{2004a}]{chai:04b}
Chai, J., Prasov, Z., \BBA\ Hong, P. \BBOP2004a\BBCP.
\newblock \BBOQ Performance evaluation and error analysis for multimodal
  reference resolution in a conversational system\BBCQ\
\newblock In {\Bem Proceedings of HLT-NAACL 2004 (Companion Volumn)}, \BPGS\
  41--44.

\bibitem[\protect\BCAY{Chai, Hong,\ \BBA\ Zhou}{Chai et~al.}{2004b}]{chai:04a}
Chai, J.~Y., Hong, P., \BBA\ Zhou, M.~X. \BBOP2004b\BBCP.
\newblock \BBOQ A probabilistic approach to reference resolution in multimodal
  user interfaces\BBCQ\
\newblock In {\Bem Proceedings of 9th International Conference on Intelligent
  User Interfaces (IUI)}, \BPGS\ 70--77.

\bibitem[\protect\BCAY{Chen, Harper,\ \BBA\ Quek}{Chen et~al.}{2002}]{chen:02}
Chen, L., Harper, M., \BBA\ Quek, F. \BBOP2002\BBCP.
\newblock \BBOQ Gesture patterns during speech repairs\BBCQ\
\newblock In {\Bem Proceedings of International Conference on Multimodal
  Interfaces (ICMI)}, \BPGS\ 155--160.

\bibitem[\protect\BCAY{Cohen}{Cohen}{1984}]{cohen:84}
Cohen, P. \BBOP1984\BBCP.
\newblock \BBOQ The pragmatics of referring and modality of communication\BBCQ\
\newblock {\Bem Computational Linguistics}, {\Bem 10}, 97--146.

\bibitem[\protect\BCAY{Cohen, Johnston, McGee, Oviatt, Pittman, Smith, Chen,\
  \BBA\ Clow}{Cohen et~al.}{1996}]{cohen:96}
Cohen, P., Johnston, M., McGee, D., Oviatt, S., Pittman, J., Smith, I., Chen,
  L., \BBA\ Clow, J. \BBOP1996\BBCP.
\newblock \BBOQ Quickset: Multimodal interaction for distributed
  applications\BBCQ\
\newblock In {\Bem Proceedings of ACM Multimedia}, \BPGS\ 31--40.

\bibitem[\protect\BCAY{Eckert\ \BBA\ Strube}{Eckert\ \BBA\
  Strube}{2000}]{eckert:00}
Eckert, M.\BBACOMMA\  \BBA\ Strube, M. \BBOP2000\BBCP.
\newblock \BBOQ Dialogue acts, synchronising units and anaphora
  resolution\BBCQ\
\newblock In {\Bem Journal of Semantics}, \lowercase{\BVOL}\ 17(1), \BPGS\
  51--89.

\bibitem[\protect\BCAY{Gold\ \BBA\ Rangarajan}{Gold\ \BBA\
  Rangarajan}{1996}]{gold:96}
Gold, S.\BBACOMMA\  \BBA\ Rangarajan, A. \BBOP1996\BBCP.
\newblock \BBOQ A graduated assignment algorithm for graph-matching\BBCQ\
\newblock {\Bem IEEE Trans. Pattern Analysis and Machine Intelligence}, {\Bem
  18\/}(4), 377--388.

\bibitem[\protect\BCAY{Gorniak\ \BBA\ Roy}{Gorniak\ \BBA\
  Roy}{2004}]{gorniak:04}
Gorniak, P.\BBACOMMA\  \BBA\ Roy, D. \BBOP2004\BBCP.
\newblock \BBOQ Grounded semantic composition for visual scenes\BBCQ\
\newblock {\Bem Journal of Artificial Intelligence Research}, {\Bem 21},
  429--470.

\bibitem[\protect\BCAY{Grice}{Grice}{1975}]{grice:75}
Grice, H.~P. \BBOP1975\BBCP.
\newblock \BBOQ Logic and conversation\BBCQ\
\newblock In Cole, P.\BBACOMMA\  \BBA\ Morgan, J.\BEDS, {\Bem Speech Acts},
  \BPGS\ 41--58. New York: Academic Press.

\bibitem[\protect\BCAY{Grosz\ \BBA\ Sidner}{Grosz\ \BBA\
  Sidner}{1986}]{grosz:86}
Grosz, B.~J.\BBACOMMA\  \BBA\ Sidner, C. \BBOP1986\BBCP.
\newblock \BBOQ Attention, intention, and the structure of discourse\BBCQ\
\newblock {\Bem Computational Linguistics}, {\Bem 12\/}(3), 175--204.

\bibitem[\protect\BCAY{Gundel, Hedberg,\ \BBA\ Zacharski}{Gundel
  et~al.}{1993}]{gundel:93}
Gundel, J.~K., Hedberg, N., \BBA\ Zacharski, R. \BBOP1993\BBCP.
\newblock \BBOQ Cognitive status and the form of referring expressions in
  discourse\BBCQ\
\newblock {\Bem Language}, {\Bem 69\/}(2), 274--307.

\bibitem[\protect\BCAY{Gustafson, Bell, Beskow, Boye, Carlson, Edlund,
  Granstrom, House,\ \BBA\ Wiren}{Gustafson et~al.}{2000}]{gustafson:00}
Gustafson, J., Bell, L., Beskow, J., Boye, J., Carlson, R., Edlund, J.,
  Granstrom, B., House, D., \BBA\ Wiren, M. \BBOP2000\BBCP.
\newblock \BBOQ Adapt - a multimodal conversational dialogue system in an
  apartment domain\BBCQ\
\newblock In {\Bem Proceedings of 6th International Conference on Spoken
  Language Processing (ICSLP)}, \lowercase{\BVOL}~2, \BPGS\ 134--137.

\bibitem[\protect\BCAY{Huls, Bos,\ \BBA\ Classen}{Huls et~al.}{1995}]{huls:95}
Huls, C., Bos, E., \BBA\ Classen, W. \BBOP1995\BBCP.
\newblock \BBOQ Automatic referent resolution of deictic and anaphoric
  expressions\BBCQ\
\newblock {\Bem Computational Linguistics}, {\Bem 21\/}(1), 59--79.

\bibitem[\protect\BCAY{Johnston}{Johnston}{1998}]{johnston:98}
Johnston, M. \BBOP1998\BBCP.
\newblock \BBOQ Unification-based multimodal parsing\BBCQ\
\newblock In {\Bem Proceedings of COLING-ACL'98}, \BPGS\ 624--630.

\bibitem[\protect\BCAY{Johnston\ \BBA\ Bangalore}{Johnston\ \BBA\
  Bangalore}{2000}]{johnston:00}
Johnston, M.\BBACOMMA\  \BBA\ Bangalore, S. \BBOP2000\BBCP.
\newblock \BBOQ Finite-state multimodal parsing and understanding\BBCQ\
\newblock In {\Bem Proceedings of COLING'00}, \BPGS\ 369--375.

\bibitem[\protect\BCAY{Johnston, Cohen, McGee, Oviatt, Pittman,\ \BBA\
  Smith}{Johnston et~al.}{1997}]{johnston:97}
Johnston, M., Cohen, P., McGee, D., Oviatt, S., Pittman, J., \BBA\ Smith, I.
  \BBOP1997\BBCP.
\newblock \BBOQ Unification-based multimodal integration\BBCQ\
\newblock In {\Bem Proceedings of ACL'97}, \BPGS\ 281--288.

\bibitem[\protect\BCAY{Kehler}{Kehler}{2000}]{kehler:00}
Kehler, A. \BBOP2000\BBCP.
\newblock \BBOQ Cognitive status and form of reference in multimodal
  human-computer interaction\BBCQ\
\newblock In {\Bem Proceedings of AAAI'00}, \BPGS\ 685--689.

\bibitem[\protect\BCAY{Koons, Sparrell,\ \BBA\ Thorisson}{Koons
  et~al.}{1993}]{koons:93}
Koons, D.~B., Sparrell, C.~J., \BBA\ Thorisson, K.~R. \BBOP1993\BBCP.
\newblock \BBOQ Integrating simultaneous input from speech, gaze, and hand
  gestures\BBCQ\
\newblock In Maybury, M.\BED, {\Bem Intelligent Multimedia Interfaces}, \BPGS\
  257--276. MIT Press.

\bibitem[\protect\BCAY{Neal\ \BBA\ Shapiro}{Neal\ \BBA\
  Shapiro}{1991}]{neal:91}
Neal, J.~G.\BBACOMMA\  \BBA\ Shapiro, S.~C. \BBOP1991\BBCP.
\newblock \BBOQ Intelligent multimedia interface technology\BBCQ\
\newblock In Sullivan, J.\BBACOMMA\  \BBA\ Tyler, S.\BEDS, {\Bem Intelligent
  User Interfaces}, \BPGS\ 45--68. ACM: New York.

\bibitem[\protect\BCAY{Neal, Thielman, Dobes, M.,\ \BBA\ Shapiro}{Neal
  et~al.}{1998}]{neal:98}
Neal, J.~G., Thielman, C.~Y., Dobes, Z.~H., M., S., \BBA\ Shapiro, S.~C.
  \BBOP1998\BBCP.
\newblock \BBOQ Natural language with integrated deictic and graphic
  gestures\BBCQ\
\newblock In Maybury, M.\BBACOMMA\  \BBA\ Wahlster, W.\BEDS, {\Bem Intelligent
  User Interfaces}, \BPGS\ 38--51. CA: Morgan Kaufmann Press.

\bibitem[\protect\BCAY{Oviatt, Coulston, Tomko, Xiao, Bunsford, Wesson,\ \BBA\
  Carmichael}{Oviatt et~al.}{2003}]{oviatt:03}
Oviatt, S., Coulston, R., Tomko, S., Xiao, B., Bunsford, R., Wesson, M., \BBA\
  Carmichael, L. \BBOP2003\BBCP.
\newblock \BBOQ Toward a theory of organized multimodal integration patterns
  during human-computer interaction\BBCQ\
\newblock In {\Bem Proceedings of Fifth International Conference on Multimodal
  Interfaces}, \BPGS\ 44--51.

\bibitem[\protect\BCAY{Oviatt, DeAngeli,\ \BBA\ Kuhn}{Oviatt
  et~al.}{1997}]{oviatt:97}
Oviatt, S., DeAngeli, A., \BBA\ Kuhn, K. \BBOP1997\BBCP.
\newblock \BBOQ Integration and synchronization of input modes during
  multimodal human-computer interaction\BBCQ\
\newblock In {\Bem Proceedings of Conference on Human Factors in Computing
  Systems: CHI'97}, \BPGS\ 415--422.

\bibitem[\protect\BCAY{Oviatt}{Oviatt}{1996}]{oviatt:96}
Oviatt, S.~L. \BBOP1996\BBCP.
\newblock \BBOQ Multimodal interfaces for dynamic interactive maps\BBCQ\
\newblock In {\Bem Proceedings of Conference on Human Factors in Computing
  Systems: CHI'96}, \BPGS\ 95--102.

\bibitem[\protect\BCAY{Oviatt}{Oviatt}{1999a}]{oviatt:00}
Oviatt, S.~L. \BBOP1999a\BBCP.
\newblock \BBOQ Multimodal system processing in mobile environments\BBCQ\
\newblock In {\Bem Proceedings of the Thirteenth Annual ACM Symposium on User
  Interface Software Technology (UIST'2000)}, \BPGS\ 21--30.

\bibitem[\protect\BCAY{Oviatt}{Oviatt}{1999b}]{oviatt:99}
Oviatt, S.~L. \BBOP1999b\BBCP.
\newblock \BBOQ Mutual disambiguation of recognition errors in a multimodal
  architecture\BBCQ\
\newblock In {\Bem Proceedings of Conference on Human Factors in Computing
  Systems: CHI'99}, \BPGS\ 576--583.

\bibitem[\protect\BCAY{Stent, Dowding, Gawron, Bratt,\ \BBA\ Moore}{Stent
  et~al.}{1999}]{stent:99}
Stent, A., Dowding, J., Gawron, J.~M., Bratt, E.~O., \BBA\ Moore, R.
  \BBOP1999\BBCP.
\newblock \BBOQ The commandtalk spoken dialog system\BBCQ\
\newblock In {\Bem Proceedings of ACL'99}, \BPGS\ 183--190.

\bibitem[\protect\BCAY{Stock}{Stock}{1993}]{stock:93}
Stock, O. \BBOP1993\BBCP.
\newblock \BBOQ Alfresco: Enjoying the combination of natural language
  processing and hypermedia for information exploration\BBCQ\
\newblock In Maybury, M.\BED, {\Bem Intelligent Multimedia Interfaces}, \BPGS\
  197--224. MIT Press.

\bibitem[\protect\BCAY{Tsai\ \BBA\ Fu}{Tsai\ \BBA\ Fu}{1979}]{tsai:79}
Tsai, W.~H.\BBACOMMA\  \BBA\ Fu, K.~S. \BBOP1979\BBCP.
\newblock \BBOQ Error-correcting isomorphism of attributed relational graphs
  for pattern analysis\BBCQ\
\newblock {\Bem IEEE Trans. Sys., Man and Cyb.}, {\Bem 9}, 757--768.

\bibitem[\protect\BCAY{Wahlster}{Wahlster}{1998}]{wahlster:98}
Wahlster, W. \BBOP1998\BBCP.
\newblock \BBOQ User and discourse models for multimodal communication\BBCQ\
\newblock In Maybury, M.\BBACOMMA\  \BBA\ Wahlster, W.\BEDS, {\Bem Intelligent
  User Interfaces}, \BPGS\ 359--370. ACM Press.

\bibitem[\protect\BCAY{Wu\ \BBA\ Oviatt}{Wu\ \BBA\ Oviatt}{1999}]{wu:99}
Wu, L.\BBACOMMA\  \BBA\ Oviatt, S. \BBOP1999\BBCP.
\newblock \BBOQ Multimodal integration - a statistical view\BBCQ\
\newblock {\Bem IEEE Transactions on Multimedia}, {\Bem 1\/}(4), 334--341.

\bibitem[\protect\BCAY{Zancanaro, Stock,\ \BBA\ Strapparava}{Zancanaro
  et~al.}{1997}]{zancanaro:97}
Zancanaro, M., Stock, O., \BBA\ Strapparava, C. \BBOP1997\BBCP.
\newblock \BBOQ Multimodal interaction for information access: Exploiting
  cohesion\BBCQ\
\newblock {\Bem Computational Intelligence}, {\Bem 13\/}(7), 439--464.

\end{thebibliography}
\bibliographystyle{theapa}

\end{document}